\documentclass{article}
\usepackage{spconf,amsmath,graphicx}
\usepackage{soul,xcolor}
\usepackage{amsfonts}
\usepackage{booktabs}
\usepackage{siunitx}
\usepackage{multirow}
\usepackage{xcolor}
\usepackage{bm}
\usepackage{url}
\usepackage[capitalize]{cleveref}
\usepackage{appendix}

\newcommand\datatwofont[1]{{\usefont{T1}{cinzeldecorative}{m}{n}#1}}
\newcommand\datatwofonttitle[1]{{{\usefont{T1}{cinzeldecorativebold}{m}{n}#1}}}

\newcommand{\benchmark}{{\datatwofont{SURE}}}
\newcommand{\newdatasetnamelong}{\datatwofonttitle{S}peech \datatwofonttitle{U}nde\datatwofonttitle{R}standing \datatwofonttitle{E}valuation}
\newcommand{\newdatasetnametitle}{{\datatwofonttitle{SURE}}}

\newcommand{\wv}{Wav2Vec 2.0}
\newlength\mylength
\title{Evaluating Parameter-efficient Transfer Learning Approaches on \newdatasetnametitle{} Benchmark for Speech Understanding}

\name{\begin{tabular}{c} Yingting Li$^{1,2}$ \qquad Ambuj Mehrish$^2$ \qquad Rishabh Bhardwaj$^2$ \qquad Navonil Majumder$^2$ \qquad Bo Cheng$^1$\\
\qquad Shuai Zhao$^1$ \qquad Amir Zadeh$^4$ \qquad Rada Mihalcea$^3$ \qquad Soujanya Poria$^2$ \end{tabular}}
\address{$^{1}$ Beijing University of Posts and Telecommunications, China \quad
$^{3}$ University of Michigan, USA \quad\\
$^{2}$ Singapore University of Technology and Design, Singapore \quad
$^{4}$ Amazon Science, USA}

\begin{document}
%
\maketitle
\begin{abstract}
Fine-tuning is widely used as the default algorithm for transfer learning from pre-trained models. Parameter inefficiency can however arise when, during transfer learning, all the parameters of a large pre-trained model need to be updated for individual downstream tasks. As the number of parameters grows, fine-tuning is prone to overfitting and catastrophic forgetting. In addition, full fine-tuning can become prohibitively expensive when the model is used for many tasks. To mitigate this issue, parameter-efficient transfer learning algorithms, such as adapters and prefix tuning, have been proposed as a way to introduce a few trainable parameters that can be plugged into large pre-trained language models such as BERT, and HuBERT. In this paper, we introduce the \newdatasetnamelong{} (\benchmark{}) benchmark for parameter-efficient learning for various speech-processing tasks. Additionally, we introduce a new adapter, ConvAdapter, based on 1D convolution. We show that ConvAdapter outperforms the standard adapters while showing comparable performance against prefix tuning and LoRA with only $0.94\%$ of trainable parameters on some of the tasks in \newdatasetnametitle{}. We further explore the effectiveness of parameter efficient transfer learning for speech synthesis task such as Text-to-Speech (TTS).
\end{abstract}
\section{Introduction}
\label{sec:intro}
Large Pre-trained Language Models (PLMs) for Natural Language Processing (NLP) are the prevalent paradigm, achieving state-of-the-art results in various language understanding tasks. 
While task-specific fine-tuning of PLM has shown remarkable success, it requires large computing resources for training as well as dedicated graphical memory due to a large number of model parameters. Moreover, fine-tuning is prone to memorization (i.e., overfitting) and catastrophic forgetting in settings with limited availability of task-specific data~\cite{tan2018survey}.
\begin{figure}
    \centering
    \includegraphics[width=0.4\textwidth]{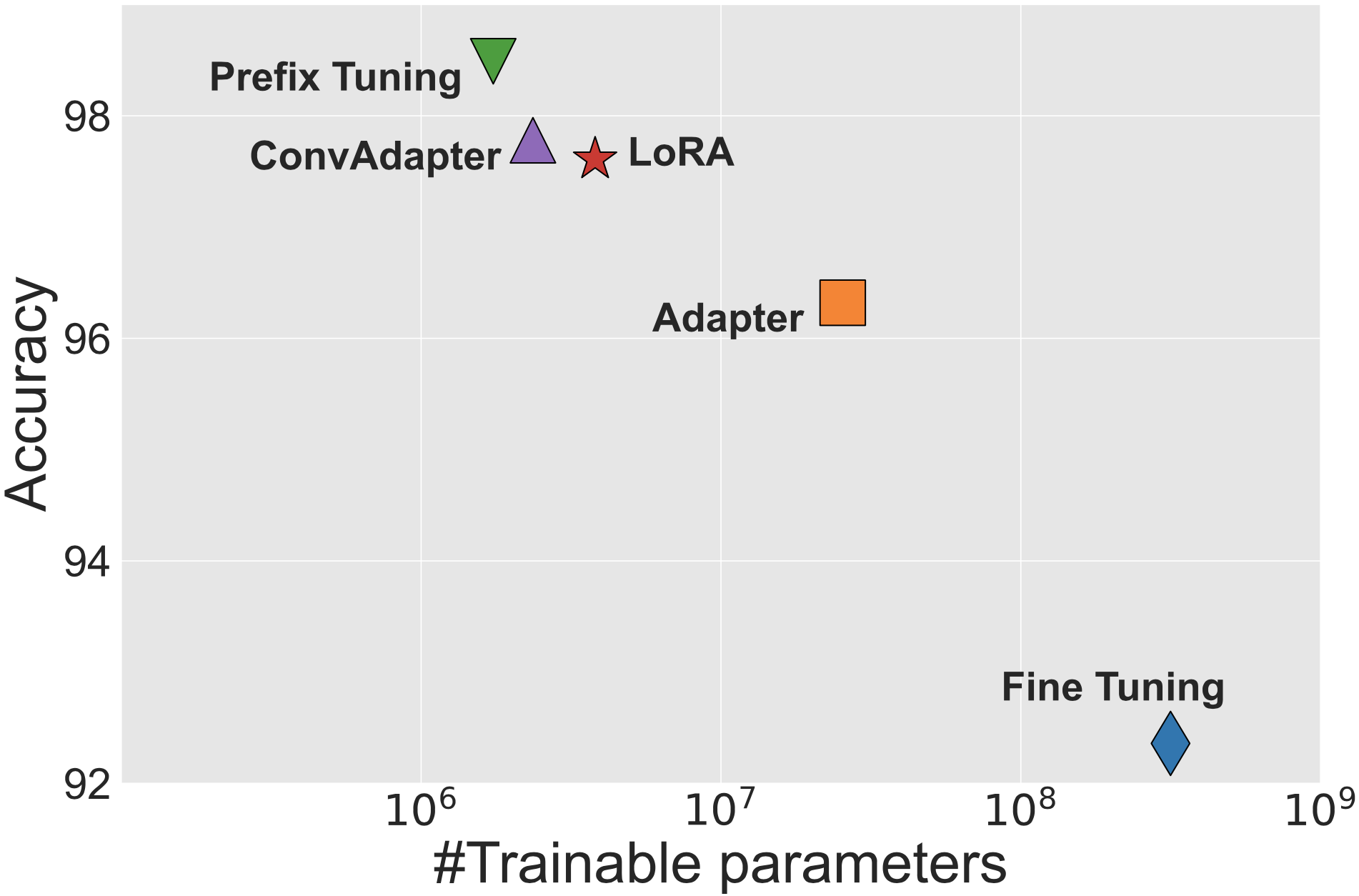}
    \caption{Accuracy (in \%) vs Trainable parameter comparison among different parameter-efficient approaches and full fine-tuning for SR task on VCTK dataset.} 
    \label{fig:param}
\end{figure}
In NLP, recent works have proposed a variety of parameter-efficient transfer-learning methods such as adapter tuning \cite{rebuffi2017learning}, prefix tuning \cite{li2021prefix}, Low-Rank Adaptation (LoRA) \cite{hu2021lora}, and prompt tuning \cite{lester2021power}, that can achieve state-of-the-art results by training only a fraction of model parameters while avoiding problems associated with full model fine-tuning. 
Keeping the large set of PLM parameters across the downstream task frozen, parameter-efficient approaches aim to train only a small set of additional task-specific parameters, thus mitigating catastrophic forgetting \cite{vander2022using}, obviating the requirement of additional memory and computing.
\begin{figure*}[t]
    \centering
    \includegraphics[width=2\columnwidth]{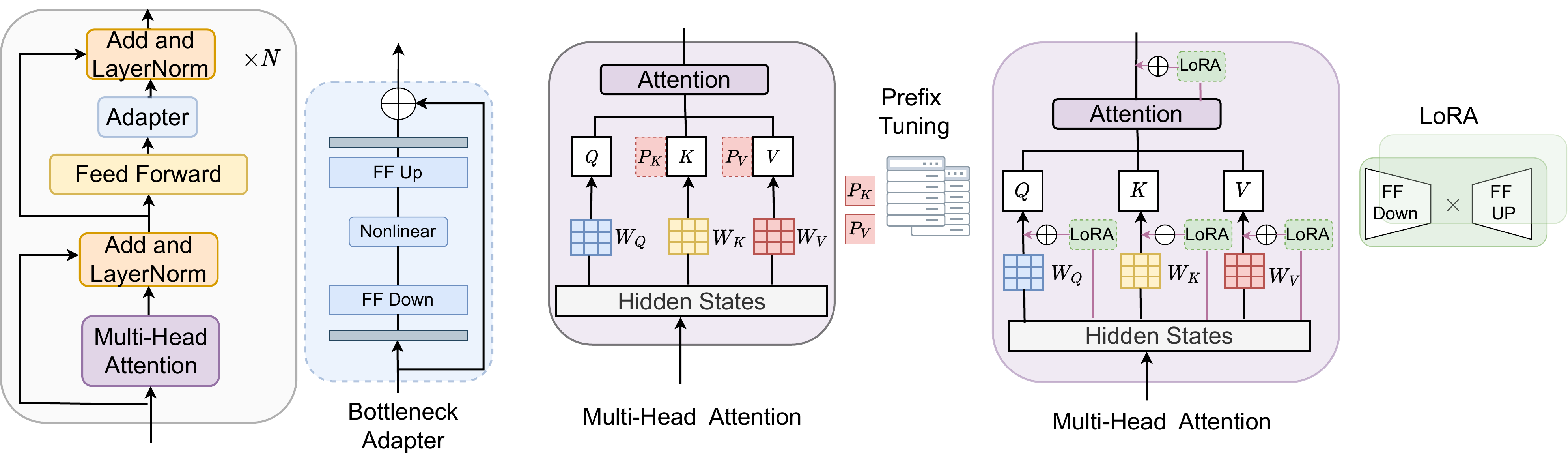}
    \caption{Transformer architecture along with Adapter, Prefix Tuning, and LoRA.}
    \label{fig:adaptarchi}
\end{figure*}
In contrast to NLP and Computer Vision (CV), the application of adapters in the speech domain has thus far been limited to a few works pertaining to Automatic Speech Recognition (ASR) \cite{thomas2022efficient}, cross-lingual/multilingual ASR \cite{kannan2019large} and Speech Translation \cite{li2021multilingual}. 
Given the irrefutable success of parameter-efficient approaches in various downstream NLP tasks, it is imperative to evaluate their efficacy in solving downstream speech processing tasks. Since the approaches to solving NLP and speech processing tasks share many characteristics, we hypothesize that the parameter-efficient algorithms in NLP can yield similar gains in speech processing.

Large PLMs in speech processing, such as \wv{}~\cite{baevski2020wav2vec}, HuBERT \cite{hsu2021hubert}, WaveLM \cite{chen2022wavlm} have evolved to encode effective representations of speech signals, reflecting in the state-of-the-art performances in downstream speech processing tasks~\cite{baevski2020effectiveness,lai2021semi}. Despite their good progress, fully fine-tuning these models for different downstream tasks comes with multiple challenges---i) requirement of a large-scale task-specific dataset, and ii) fine-tuning all the model parameters and allocating a dedicated memory for each downstream task. For instance, in sentiment analysis, training data often have low data volume with noisy labels owing to the time and cost required for data collection and annotation. Furthermore, the largest version of \wv{}, contains 317 million parameters. Since the challenges associated with fine-tuning NLP and speech-based PLMs are alike, we posit that parameter-efficient approaches from NLP can benefit speech-processing downstream applications. Thus, our contributions are:

\noindent \textbf{[Contribution 1]} We provide an effective setup of state-of-the-art parameter-efficient approaches for different speech processing tasks such as Speech Emotion Recognition (SER), Speaker Recognition (SR), ASR, and Keyword Spotting (KS). For this, we introduce a new benchmark \benchmark{} to examine the capability of different parameter-efficient approaches, i.e., Adapter tuning, Prefix tuning, and LoRA. 

In the speech domain, the strong correlations between adjacent time-frequency components of speech signals can be exploited further by using Convolutional neural networks (CNNs). This motivates our second contribution.

\noindent \textbf{[Contribution 2]} We introduce a new lightweight adapter, ConvAdapter, based on CNNs. While in this work we use \wv, ConvAdapter can scale well to any transformer-based speech model, including HuBERT \cite{hsu2021hubert}. We hope the results and analyses presented in this work will further motivate the development of more generalizable and reusable parameter-efficient approaches to advance speech research.
%
%
%
\section{Related Work}
\label{sec:relatedworks}
To explore the capability of Large PLMs in various speech processing tasks, Yang et al \cite{yang2021superb} introduce the SUPERB benchmark. They freeze the pre-trained self-supervised learning (SSL) model and introduce a task-specific prediction layer on top of the model. This is an approach of the addition of task-relevant description to the downstream input to help the pre-trained model to understand the downstream tasks. However, it fails to explore the capability of the upstream model.

Adapters are introduced in NLP as parameter efficient algorithms for transfer learning \cite{houlsby2019parameter}. In various NLP tasks, adapters have shown remarkable success in achieving on-par performance with full fine-tuning with a significantly smaller number of trainable parameters. The advantage of adapters has also been observed in CV tasks \cite{rebuffi2017learning}.
The use of adapters in speech processing is mostly limited to ASR \cite{thomas2022efficient,zhu2020multilingual}.  Recently, they have also been explored for speech translation as well, but in a limited scope. \cite{escolano2021enabling} addressed a very specific setting (zero-shot ST), while \cite{li2021multilingual} used only a single adapter after a transformer encoder. Recurrent Neural Network Transducer (RNN-T) \cite{kannan2019large}  architecture introduces an adapter module for a streaming end-to-end multilingual system. The effectiveness of adapters in continual learning is outlined in \cite{kessler2022adapter}, where authors report a significant decrease in pre-training time of the model on new language tasks without forgetting previous representation. Additionally, the authors in \cite{chang2022exploration} explore the use of speech prompts for the Generative Spoken Language Model \cite{lakhotia2021generative}.


\section{Parameter-efficient Transfer Learning}
\begin{figure}[h]
    \centering
    \includegraphics[width=\columnwidth]{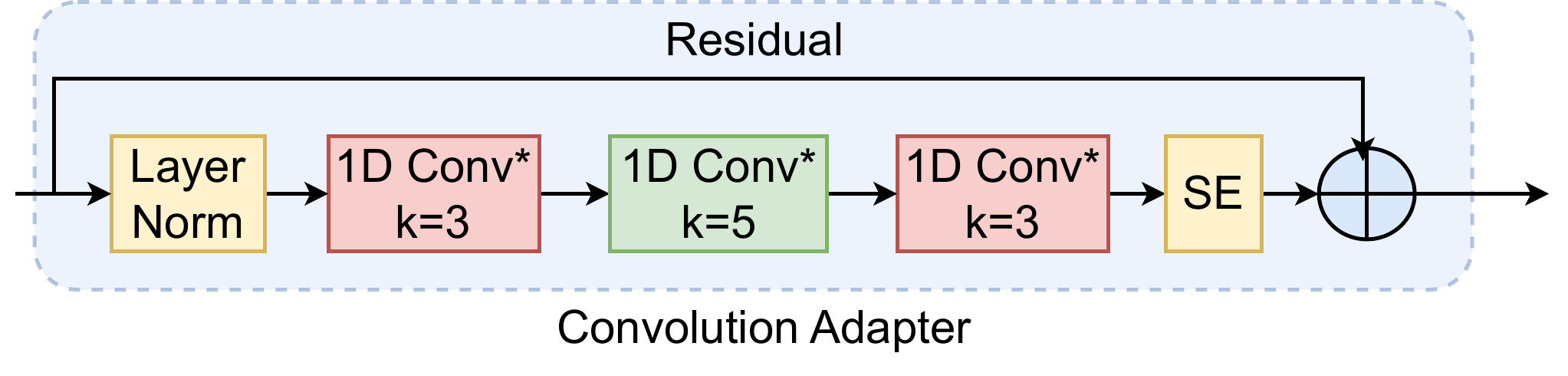}
    \caption{The architecture of 1D convolution layer-based lightweight adapter. $k$ is the kernel size of 1D convolution. $*$ denotes depth-wise convolution.}
    \label{fig:convadapt}
\end{figure}

Next, we elaborate on the different approaches considered for parameter-efficient transfer learning. \Cref{fig:adaptarchi} and \Cref{fig:convadapt} illustrate the different approaches.

\vspace{1em}
\noindent \textbf{Adapter Tuning.} A (bottleneck or standard) adapter \cite{houlsby2019parameter, pfeiffer2020adapterfusion} is a neural module with a significantly small number of parameters as compared to LM. It down-projects the input $h \in \mathbf{R}^{d}$  to a $m$-dimensional space ($m<d$), applies a non-linearity $g(\cdot)$, and finally up-projects the result back to $d$-dimensions. A residual connection is added to obtain the final output as
\begin{equation}
    \bm{h} \leftarrow \bm{h} + g(\bm{h} \bm{W}_{\text{down}}) \bm{W}_{\text{up}}
\end{equation}
where $\bm{W}_{\text{down}} \in \mathbb{R}^{d \times m}$ and $\bm{W}_{\text{up}} \in \mathbb{R}^{m \times d}$ are down and up projection matrices, respectively. It has been found empirically that a two-layer feed-forward neural network with a bottleneck works well. 
While there are many adapter variants, in our experiments, we adopt the settings from \cite{pfeiffer2020adapterfusion}. As shown in \Cref{fig:adaptarchi}, the adapter is inserted after the feed-forward layer of every transformer module.

\vspace{1em}
\noindent \textbf{Prefix tuning.} It changes the attention module of the Transformer by prepending $k$ learnable vectors to the pre-trained multi-head attention keys and values at every layer. As illustrated in \Cref{fig:adaptarchi}, two sets of learnable prefix vectors $\bm{P}_{K}$ and $\bm{P}_{V}$ are concatenated with the original key $\bm{K}$ and value $\bm{V}$ and query $\bm{Q}$ remains unchanged. The new keys and values are then used for multi-head attention. The $\text{head}_i$ of multi-head attention is given as:
\begin{equation}
\text{head}_{i} = \text{Attn}(\bm{Q}\bm{W}_{Q}^{(i)},[\bm{P}_{K}^{(i)},\bm{K}\bm{W}_{Q}^{(i)}],[\bm{P}_{V}^{(i)},\bm{V}\bm{W}_{Q}^{(i)}]) 
\end{equation}
where Attn($\cdot$) is scaled dot-product attention given by:
\begin{equation}
\label{attn}
    \text{Attn}(\bm{Q},\bm{K},\bm{V}) = \text{softmax} (\frac{\bm{Q}\bm{K}^{T}}{\sqrt{d_{k}}})\bm{V}
\end{equation}
During training, only prefix vectors  $\bm{P}_{K}$ and $\bm{P}_{V}$ are updated. Since prefix tuning modifies
the attention head in each layer, it is likely to have more control of acoustic information being passed from one layer to another and hence can effectively stimulate the knowledge of the pre-trained model.

\vspace{1em}
\noindent \textbf{LoRA.} Proposed by \cite{hu2021lora}, it approximates the weight updates by injecting trainable low-rank matrices into transformer layers. The weight update for pre-trained weight matrix $W \in \mathbb{R}^{d \times k}$ is represented by low-rank decomposition $\bm{W} + \Delta \bm{W} = \bm{W} + \bm{W}_{\text{down}} \bm{W}_{\text{up}}$ where $\bm{W}_{\text{down}} \in \mathbb{R}^{d \times r}$, $\bm{W}_{\text{up}} \in \mathbb{R}^{r \times k}$ are tunable parameters and $r$ is the rank of the decomposition matrices with $r<d$. For specific input $\bm{x}$ to the linear projection in the multi-headed attention layer, LoRA modifies the projection output $\bm{h}$ as
\begin{equation}
\label{lora}
    \bm{h} \leftarrow \bm{h} + s\cdot \bm{x} \bm{W}_{\text{down}}\bm{W}_{\text{up}}
\end{equation}
In this work, we add LoRA at four places in the multi-head attention layer, as shown in \Cref{fig:adaptarchi}. Due to the significantly lightweight, many small modules can be added to the pre-trained model for different tasks and can effectively switch tasks by replacing the modules. Also, compared with the fully fine-tuned model, LoRA introduces no inference latency and roughly converges to training the original model \cite{hu2021lora}.

\noindent \textbf{Convolutional Adapter (Ours).} CNNs are successful and valuable models for working with speech processing systems. Many studies in the past have validated the effectiveness of CNN for speech. It can fuse channel-wise information together within the local receptive field while learning task-specific information. Hence, can be effective in  distilling the knowledge encoded in pre-trained model parameters. Inspired by this property, we introduce a lightweight adapter based on CNN layers to learn task-specific representations. We implement the adapter using three lightweight 1D convolutional layers with kernel size $k$ as $3$ and $5$ (convolution layer with $k = 5$ is  depth-wise convolution \cite{chollet2017xception})  along with layer norm and squeeze and excite module\cite{https://doi.org/10.48550/arxiv.1709.01507}. \cref{fig:convadapt} illustrate the architecture of the proposed adapter. This adapter is referred to as \emph{ConvAdapter} in the rest of the paper. We use depth-wise convolution since it requires fewer parameters than standard convolution and is computationally more efficient. Finally, the output is obtained by adding the residual connection. ConvAdapter is inserted at the same location as Bottleneck Adapter as shown in \cref{fig:adaptarchi}.
\section{Experiments and Results}
\subsection{The \benchmark{} Benchmark}
To evaluate the methods, we set up a benchmark of four different speech-processing tasks with diverse datasets, called \newdatasetnamelong{} (\benchmark{}). A brief description of each task, along with the datasets in \benchmark{}, is shown in \cref{tab:dataset}. As an evaluation metric, accuracy is used for SER, SR, IC, and KS and word error rate and (wer$\%$) and phoneme error rate (per$\%$) is used for ASR and PR respectively. Speech Language Understanding (SLU) tasks such as SF  requires both slot-types and slot-values \cite{lai2021semi}. Hence, slot-type F1 score and slot value CER \cite{lugosch2019speech} are used as evaluation metrics. TTS is results are presented for both objective and subjective evaluation (the details of the metric used for evaluation is discussed in Section \ref{tts})
\begin{table*}[h]
    \centering
    \begin{tabular}{@{}llll@{}}
    \toprule
    {Tasks} &{Type}& {Dataset} & {$N$}\\
    \midrule
    {Speech Emotion Recognition (SER)}& Recognition &ESD \cite{zhu2020multilingual}, MELD\cite{poria2019meld}&5, 7\\
    {Speaker Recognition (SR)} & Recognition &ESD \cite{zhu2020multilingual}, VCTK\cite{veaux2017cstr}&10, 110\\
    {Automatic Speech Recognition (ASR)} & Generation &ESD \cite{zhu2020multilingual}, FLEURS \cite{conneau2022fleurs}, LS \cite{panayotov2015librispeech}&33, 37\\
    {Phoneme Recognition (PR)} & Recognition & LS \cite{panayotov2015librispeech}&11\\
    {Intent Classification (IC)} & Spoken Language Understanding &Fluent Speech Commands\cite{lugosch2019speech}&5\\
    {Slot Filling (SF)} & Spoken Language Understanding&SNIPS\cite{coucke2018snips}&11\\
    {Text-To-Speech (TTS)} & Generation &LTS \cite{panayotov2015librispeech}, L2ARCTIC \cite{zhao2018l2}&247,24\\
    {Keyword Spotting (KS)} & Classification &Speech Commands\cite{warden2018speech}&11\\\bottomrule
    \end{tabular}
    \caption{The tasks in the \newdatasetnametitle{} benchmark. $N$ is the number of classes (size of vocabulary for ASR) in each category. CLS: Classification, SG: Sequence Generation. For, TTS $N$ represents the number of speakers in each dataset.}
    \label{tab:dataset}
\end{table*}
\begin{table*}[h]
    \centering
    \begin{tabular}{@{}llcccccccc@{}} 
    \toprule
    \multirow{2}{*}{Method}& \multirow{2}{*}{\#Parameters} &\multicolumn{2}{c}{SER (acc \% / w-f1) $\uparrow{}$} & \multicolumn{2}{c}{SR (acc \%) $\uparrow{}$} &\multicolumn{3}{c}{ASR (wer) $\downarrow$} &\multicolumn{1}{c} {KS (acc \%) $\uparrow$}\\
    \cmidrule(lr){3-4}\cmidrule(lr){5-6}\cmidrule(lr){7-9} \cmidrule(lr){10-10}
    {}& {}&ESD  & MELD & ESD  & VCTK  & ESD& FLEURS& LS & Speech Command \\
    \midrule
    {Fine Tuning} & 315,703,947 & \textbf{96.53} & 42.93 & 99.00 & 92.36 & 0.2295 & \textbf{0.135} &\textbf{0.0903 }&99.08\\
    {Adapter}& 25,467,915 (8.08\%) & \underline{94.07}  & 41.58 & 98.87  & 96.32 & \underline{0.2290} & 0.214 & 0.2425&99.19 \\
    {Prefix Tuning} & 1,739,787 (\textbf{0.55\%})  & 90.00  & \underline{44.21} & \textbf{99.73}  & \textbf{98.49} & \textbf{0.2255} & 0.166 & 0.1022&\underline{98.86}\\
    {LoRA}& 3,804,171 (1.20\%) & 90.00  & \textbf{47.05} & 99.00  & 97.61 & 0.2428 & \underline{0.149} & \underline{0.1014}&98.28 \\
    {ConvAdapter}& 2,952,539 (\underline{0.94\%}) & 91.87  & 46.30 & \underline{99.60}  & \underline{97.61} & 0.2456 & 0.2062 & 0.2958 &\textbf{98.99} \\\bottomrule
\end{tabular}
   \caption{\newdatasetnametitle{} benchmark on full fine-tuning and other parameter-efficient training methods on pre-trained \wv{}. \#Parameters is the number of KS task's trainable parameters. The percentage in the parentheses represents the fraction of trainable parameters with respect to full fine-tuning. The best performance is in bold and the second best is underlined. Due to data imbalance, we use weighted-f1 as a metric (abbreviated as w-f1) on MELD.}
    \label{tab:performance}
\end{table*}

\subsection{Training}
We use the standard \wv{} architecture (XLSR-53-english) with $24$ transformer modules and the publicly available pre-trained checkpoint from  Hugging Face\footnote{\url{https://huggingface.co/jonatasgrosman/wav2vec2-large-xlsr-53-english}} that is fine-tuned on XLSR-53 on English using the train and validation splits of Common Voice 6.1. XLSR-53 is trained on  CommonVoice, BABLE, and  Multilingual LibriSpeech.
We add the adapters to each transformer layer, as shown in \cref{fig:adaptarchi}. Finally, a fully-connected layer is appended to the final layer of the pre-trained model for the task-specific output\footnote{Code available at \url{https://github.com/declare-lab/speech-adapters}}. For each task, we partition the datasets into three sets: train, validation, and test. While adapting to the new task due to limited data, training for too many epochs can lead to overfitting. 
 To avoid this, we employ early stopping. When no improvement in the validation set performance over five epochs, we stop the training and save the best checkpoint for evaluation over the test set.
\subsection{Results and Discussions}
\label{sec:results}
\cref{tab:performance} and \cref{tab2:performance} demonstrate that the studied parameter-efficient approaches perform on par with or better than full fine-tuning on various speech processing tasks. Their results on KS and ASR are comparable to fine-tuning. On the ESD dataset, these approaches outperform Fine-tuning by a large margin. Even for the MELD dataset, the performance is better than Fine-tuning. However, the overall performance on the MELD dataset is not good since it is a multi-modal dataset for emotion recognition, and we use only a single modality for this work. In the SR task, we observe that representation learned using these tuning paradigms can recognize the speaker effectively, even with emotional and accented speech. Moreover, the number of trainable parameters required to achieve these results is significantly lower than fine-tuning --- Adapter and ConvAdapter require only $8.08\%$ and $0.94\%$ of the trainable parameters of full fine-tuning, respectively. \Cref{fig:param} illustrates a comparative view of trainable parameter count vs performance for SR task.

The performance of the proposed ConvAdapter\footnote{ASR and SF results for Convadapter require further tuning for optimal hyperparameters.} is similar to or slightly worse than other adapters and fine-tuning. This small performance penalty, however, comes with a huge reduction in the number of trainable task-specific parameters. With respect to fine-tuning, ConvAdapter has only $0.94\%$ of the trainable parameters. ConvAdapter performs similarly to the standard adapter and LoRA with significantly fewer parameters. However, prefix tuning outperforms the ConvAdapter in both performance and trainable parameter count (only $0.55\%$ with respect to fine-Tuning). Since prefix tuning typically modifies the output of each attention head in each transformer module, it is more expressive than other adapters.

In addition to tasks outlined in \cref{tab:performance} and \cref{tab2:performance}, we also expand our experiments to speech synthesis. The following section will illustrate the effectiveness of the parameter-efficient training methods disused in \benchmark{} benchmark on Text-to-Speech (TTS).
\renewcommand{\arraystretch}{1.2}
\begin{table*}[]
\centering
\resizebox{2\columnwidth}{!}{
\begin{tabular}{llllllll}
\hline
\multicolumn{1}{c}{\multirow{3}{*}{Method}} & \multicolumn{2}{c}{IC}                                                     & \multicolumn{2}{c}{PR}                                                     & \multicolumn{3}{c}{SF}                                                                               \\ \cline{2-8} 
\multicolumn{1}{c}{}                        & \multicolumn{1}{c}{\multirow{2}{*}{\#Parameters}} & \multicolumn{1}{c}{FS} & \multicolumn{1}{c}{\multirow{2}{*}{\#Parameters}} & \multicolumn{1}{c}{LS} & \multicolumn{1}{c}{\multirow{2}{*}{\#Parameters}} & \multicolumn{2}{c}{SNIPS}                        \\ \cline{3-3} \cline{5-5} \cline{7-8} 
\multicolumn{1}{c}{}                        & \multicolumn{1}{c}{}                              & ACC\% $\uparrow$       & \multicolumn{1}{c}{}                              & PER $\downarrow$     & \multicolumn{1}{c}{}                              & \multicolumn{1}{c}{F1 \% $\uparrow$} & \multicolumn{1}{c}{CER $\downarrow$} \\ \hline
Fine-Tuning                                 & 315707288                                         & 99.60                  & 311304394                                         & 0.0577                 & 311375119                                         & 93.89                  & 0.1411                  \\
Adapter                                     & 25471256  (8.06\%)                                & 99.39                  & 25278538 (8.01\%)                                 & 0.1571                 & 25349263 (8.14\%)                                 & 92.60                   & 0.1666                  \\
Prefix Tuning                               & 1743128  (0.55\%)                                 & 93.43                  & 1550410 (0.49\%)                                  & 0.1598                 & 1621135 (0.50\%)                                  & 62.32                  & 0.6041                  \\
LoRA                                        & 3807512 (1.20\%)                                  & 99.68                  & 3614794 (1.16\%)                                  & 0.1053                 & 3685519 (1.18\%)                                  & 90.61                      & 0.2016                       \\
ConvAdapter                                 & 3672344 (1.16\%)                                  & 95.60                  & 3479626 (1.11\%)                                  & 0.1532                 & 3550351 (1.14\%)                                  & 59.27                      & 0.6405                    \\ \hline
\end{tabular}}
\caption{\newdatasetnametitle{}  benchmark on full fine-tuning and other parameter-efficient training methods on pre-trained Wav2Vec 2.0 for IC and PR tasks on  \textbf{FS}: Fluent Speech \cite{lugosch2019speech} and \textbf{LS}: LibriSpeech \cite{panayotov2015librispeech} datasets, respectively.}
\label{tab2:performance}
\end{table*}
\renewcommand{\arraystretch}{1.2}
\begin{table*}[h]
\centering
\begin{tabular}{lccccc}
\hline
\multicolumn{1}{c}{\multirow{2}{*}{Method}} & \multirow{2}{*}{Parameters (\%)} & \multicolumn{2}{c}{LTS} & \multicolumn{2}{c}{L2ARCTIC} \\ \cline{3-6} 
\multicolumn{1}{c}{}                        &                                  & MCD $\downarrow$       & WER   $\downarrow$      & MCD $\downarrow$           & WER $\downarrow$          \\ \hline
Finetuning                                  & 35802977                         & 6.2038     & 0.2655     & 6.71469       & 0.2141       \\
Adapter                                  & 659200                           & 6.1634     & 0.3143     & 6.544         & 0.2504       \\
Prefix                                      & 153600                           & 6.2523     & 0.3334     & 7.4264        & 0.3244       \\
LoRA                                        & 81920                            & 6.8319     & 0.3786     & 7.0698        & 0.3291       \\
Convadapter                                 & 108800                           & 6.9202     & 0.3365     & 6.9712        & 0.3227       \\ \hline
\end{tabular}
\caption{MCD and WER of samples generated using different methods.}
\end{table*}
\subsection{Text-To-Speech (TTS)}
\label{tts}
The goal of Text-to-Speech (TTS), also known as speech synthesis, is to synthesize intelligible and natural speech from text. Morden TTS systems consist of complex neural networks to transform the natural language modelling process into audible speech output \cite{shen2018natural,ren2019fastspeech,ren2020fastspeech}. The  advances in neural network-based TTS systems have significantly improved the quality of synthesized speech. Neural network-based TTS systems consist of three basic building modules for text analysis, acoustic modelling, and waveform generation.

We use Transformer based TTS \cite{li2019neural} as our base model. Transformer TTS utilizes Transformer \cite{vaswani2017attention} and Tacotron2 \cite{shen2018natural}  as its building block. The detail architecture of Transformer TTS is outlined in Section \ref{appendix}. In this work, we pre-train Multi-speaker Transformer TTS on VCTK \cite{veaux2017cstr} dataset. For adapting the base model using parameter efficient methods. we focus our experiments on publicly available LibriTTS \cite{panayotov2015librispeech} and L2ARCTIC \cite{zhu2020multilingual} dataset. We first train the base model for $900$k steps. During adaptation, we freeze the base model completely and only updated the parameters of additionally added modules (Adapter, Prefix, LoRA and Convadapter). We fine-tune the model for additional $40$k steps and synthesize speech for the text samples in the validation set of each speaker and evaluate the synthesized speech on the following aspects.
    \subsubsection{Objective Evaluation} Objective evaluations are performed by comparing generated samples distortion of acoustic features against ground truth reference samples. The score from these assessments are not necessarily correlated with human judgment. Mel-ceptrum Distortion (MCD) score and Word Error rate (wer) are used as metrics. MCD  is a measure of the difference between two sequences of mel cepstra, whereas WER can serve as a comparison metric for different speech synthesis models to measure the intelligibility of generated speech among them.
    \begin{itemize}
        \item Mel-ceptrum Distortion (MCD): MCD \cite{kubichek1993mel} can be computed using Mel Spectrogram of synthesized (target) and reference utterances using Equation \ref{mcd}.
        \begin{equation}
        \label{mcd}
            MCD[dB] = \frac{10}{\ln 10}\sqrt{2\sum_{i=1}^{24}(m^{t}_{k,i}-m^{r}_{k,i})^{2}}
        \end{equation}
        where $m_{k,i}^{t}$ and $m_{k,i}^{r}$, are the Mel-ceptal coefficients (MCEPs) of target and reference utterances respectively for $k^{th}$ frame.
        \item Word Error rate (wer): In our experiments, we use enterprise-grade speech-to-text (STT) pre-trained silero models \cite{wer} for computing wer. We compute WER for ground truth utterances to allow for relative comparison between the models.
    \end{itemize}
    In terms of MCD and WER score, Adapter outperforms other methods on LTS dataset, whereas Convadapter performers better than LoRA and Prefix tuning on L2ARCTIC dataset. The difference in performance can be attributed to the fact that L2ARCTIC is accented dataset consists of different accents other than native English accent, whereas LTS is mostly native English accent. The Convadapter due to it architecture is able to adapt to change in accent better than Prefix and LoRA. 
    \subsection{Subjective Evaluation} 
\begin{figure}
    \centering
    \includegraphics[width = \columnwidth]{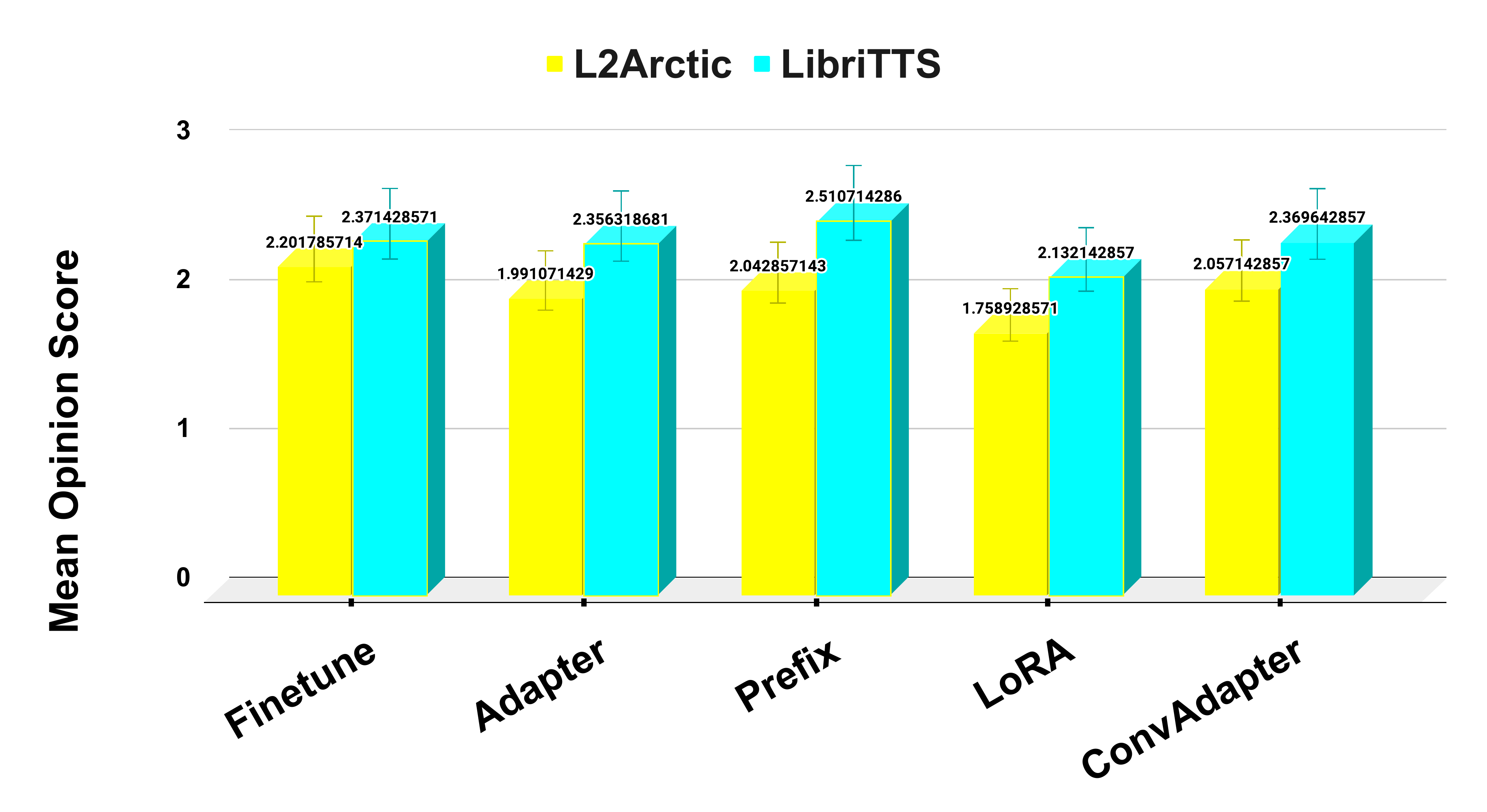}
    \caption{Speaker Similarity, different parameter efficient methods.}
    \label{spksim}
\end{figure}

To evaluate samples generated by the model subjectively, Mean Opinion Score (MOS) tests were conducted for naturalness and speaker similarity. 
    \begin{itemize}
        \item Naturalness: To perform naturalness evaluations, we synthesize speech samples of the unseen text in the validation set for all speakers and, asked listener to rate each audio utterance on a 5-point naturalness scale ($1$: bad, $2$: poor, $3$: fair, $4$: good, $5$:excellent) based on the measurement objective of the test. 
        \item Speaker Similarity: To evaluate the speaker similarity of the synthesize speech with target/reference speaker, we perform speaker similarity evaluation. We presented the ground truth speech of the target speaker along with speech samples generated by different method and, asked listener to rate each speech utterance on a 5-point speaker similarity scale (similar to naturalness) with the respect to ground truth utterance.
    \end{itemize}
\begin{figure}
    \centering
    \includegraphics[width = \columnwidth]{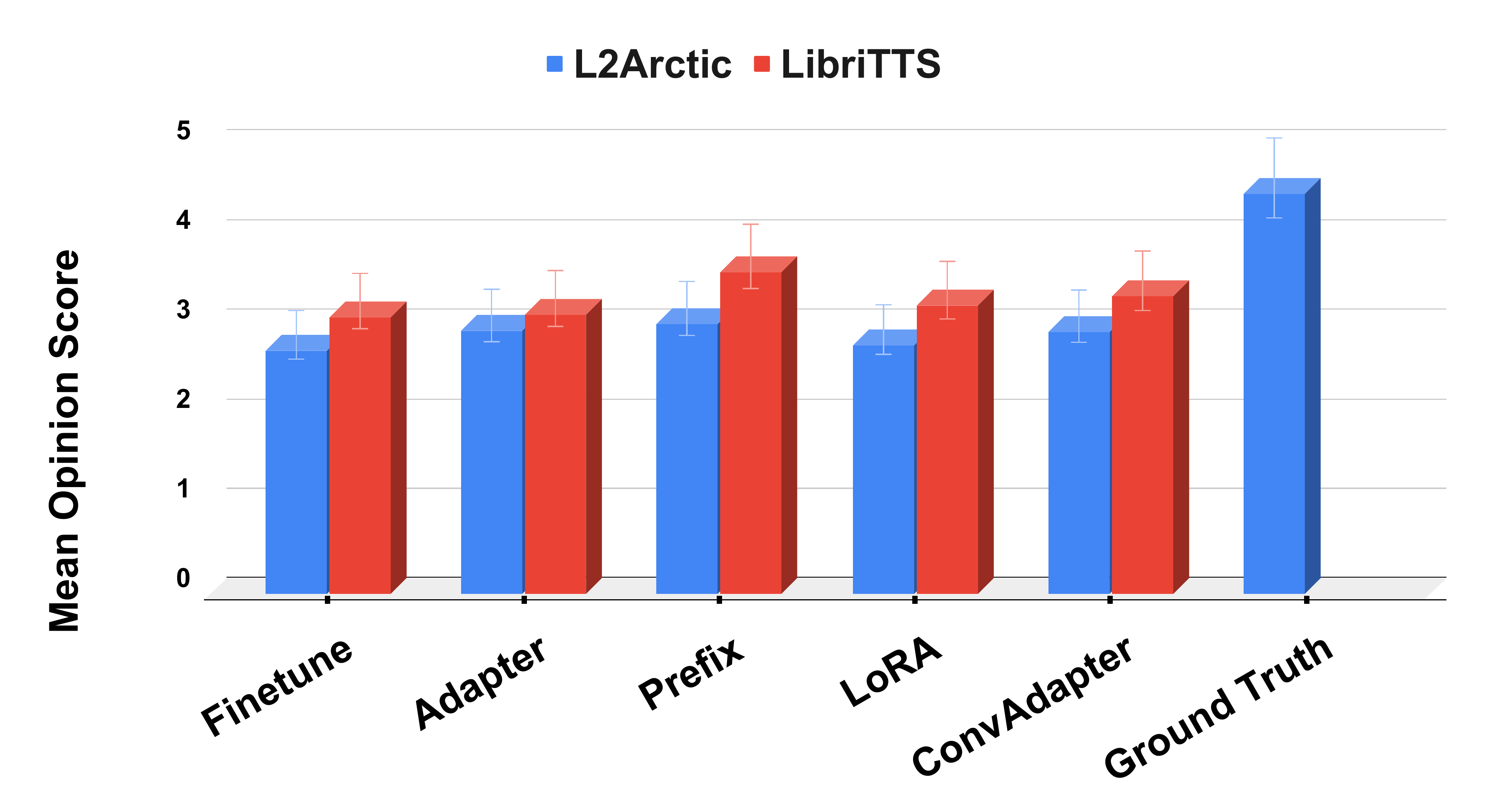}
    \caption{Naturalness for different parameter efficient methods. We present the results on validation data and all models are trained for 40k steps.}
    \label{naturalness}
\end{figure}
In total, $9$ responses were gathered from subjective evaluation from the researchers working different areas of NLP. Each participant listen to $180$ samples selected randomly from the generated samples. (For template and description of the test, please refer to appendix). We present the naturalness  and speaker similarity scores with $95\%$ confidence interval in Figure \ref{naturalness} and \ref{spksim}. Interestingly, parameter efficient approaches archives similar MOS score to full fine-tuning in both naturalness and speaker similarity. We also observe that the fine-tuning approach performs slightly better in speaker similarity.   We encourage the readers to listen to the speech samples to observe the difference between each method\footnote{\url{https://declare-lab.github.io/speechadapters/}}. 

\renewcommand{\arraystretch}{1.5}
\begin{table*}[h]
\centering
\begin{tabular}{llcccc}
\hline
Top k layers                 & k            & 24        & 16        & 8         & 4        \\ \hline
\multirow{2}{*}{Finetuning}  & \#Parameters & 315703690 & 214933898 & 114164106 & 63779210 \\
                             & Performance  & 0.9900    & 0.9967    & 0.9980    & 0.9953   \\ \hline
Compression rate             & n ($2^n$)    & 1 (2)     & 2 (4)     & 3 (8)     & 4 (16)   \\ \hline
\multirow{2}{*}{Bottleneck}  & \#Parameters & 25467658  & 12878602  & 6584074   & 3436810  \\
                             & Performance  & 0.9887    & 0.9667    & 0.9093    & 0.8313   \\ \hline
\multirow{2}{*}{Convadapter} & \#Parameters & 3668746   & 3638026   & 3622666   & 3614986  \\
                             & Performance  & 0.9960    & 0.9947    & 0.9967    & 0.9973   \\ \hline
\end{tabular}
\caption{Validation set accuracy and a number of trainable parameters for Fine-tuning, Bottleneck, and ConvAdapter. Bottleneck and ConvAdapter are tuned with size $2^n$ for $n=1,2,3,4$. In Fine-tuning top $k$ layers are fine-tuned for $k=24,16,8,4$.}
\label{topk}
\end{table*}

\subsection{Discussion}
The significant performance gain with few trainable parameters could be attributed to the fact that, while learning the task-specific adapters, the gradients can be back-propagated all the way through the pre-trained model, distilling the rich knowledge encoded in its parameters. The ConvAdapter we propose allows us to achieve better performance with a fraction of parameters, thus providing an alternative parameter-efficient transfer learning method. Since ConvAdapter requires only $0.94\%$ of the trainable parameters, it would be interesting to combine it with other adapters, such as prefix fine-tuning or LoRA to further improve the results. Especially, for new and challenging datasets like MELD \cite{poria2019meld}, where performance is subpar. The results reported in \cref{tab:performance} show that using adapters with pre-trained SSL models is a promising future research direction. For future work, we can extend the benchmark to different transformer-based speech models for tasks like Speech translation, Voice conversion, and Text-to-Speech. Also, there is much room for investigating the representations learned by these models in multilingual settings. Since \wv{}-XLSR-53 is trained on 53 different languages, it will be interesting to evaluate the cross-lingual representations learned by these models.

The previous work in parameter efficient fine-tuning discussed possible cost to performance based on size of the adapters \cite{houlsby2019parameter}. Smaller adapters introduce fewer parameters and can lead to degrade in performance.  Therefore, to study the impact of size of adapter on the per romance, we also perform a parameter/performance trade-off experiment. Following the set-up discussed in \cite{houlsby2019parameter}, we consider different adapter size and compared it with Fine-tuning of only the top $k$ layers of \wv{}. Table \ref{topk} shows the parameter/performance trade-off for the Speaker Recognition task.  We can observe from Table \ref{topk} that, the performance of Convadapter is not impacted by changing the compression rate, whereas the performance of bottleneck degrades. The results of fine-tuning are particularly interesting, since there is no performance drop. The possible reason for the same is that speaker information is encoded in initial layers of wav2vec2 model. These results highlight the parameter robust aspect of Convadapter for speech analysis. Although the performance of the Convadapter similar to other methods, but it provides robust method for fine-tuning the large pre-train model effectively for various speech processing tasks. This analysis further opens up the possibility of exploring information encoded at different layers of the large pre-trained models. This information can be exploited further to make the training more parameter efficient by only fine-tuning the specific layer for specific tasks. 

\section{Conclusion and Future work}
\label{sec:conclusion}
In this work, we introduced the \benchmark{} benchmark and explored the effectiveness of different parameter-efficient transformer-based adapters on this benchmark. We performed extensive experiments to investigate the capability of these methods on a wide range of speech-processing tasks. In our experiments, we showed that adapters with few parameters are capable of achieving comparable or even better performance compared to full fine-tuning. Finally, we proposed ConvAdapter, which can achieve state-of-the-art performance with a limited number of trainable parameters.

Although, parameter efficient methods are able to produce the results similar to fine-tuning for TTS, however there is still room for improvement. During our experiments, we notice that the performance is sup-par when we try to generate longer sentences. Also, the quality of speech samples can be further improved. In future work, we can explore training of high-quality TTS models in limited data and resource constrained settings. We hope this work will inspire future research on parameter-efficient algorithms for speech-related tasks.

\bibliographystyle{IEEEbib}
\bibliography{refs}
\clearpage
\newpage

\section{Appendix}
\label{appendix}
\subsection{Pre-trained Model:\wv{}}
\label{sec:typestyle}
\wv{} consists of three main modules: feature encoder, quantization module, and context network. A feature encoder composes of seven temporal convolutional blocks. It transforms speech input $X$ (raw mono waveform at 16 kHz) into latent representation $z$. The quantization module discretized $z$ into $q$, a dictionary of discrete and latent representations of speech. Finally, a context network consisting of $12$ transformer blocks generates contextualized embedding $c$, of the same dimensionality of $q$. The model is trained using a self-supervised paradigm optimizing two losses. The first loss is contrastive and requires the model to predict the quantized representation q of some masked input using $c$, from a finite set of quantized representations drawn from, the input sample. The second loss ensures that the quantized representations are diverse \cite{baevski2020wav2vec}.
\subsection{TTS Architecture}
\begin{figure}[h]
\large
    \centering
    \includegraphics[width = \columnwidth,height=1.5\columnwidth]{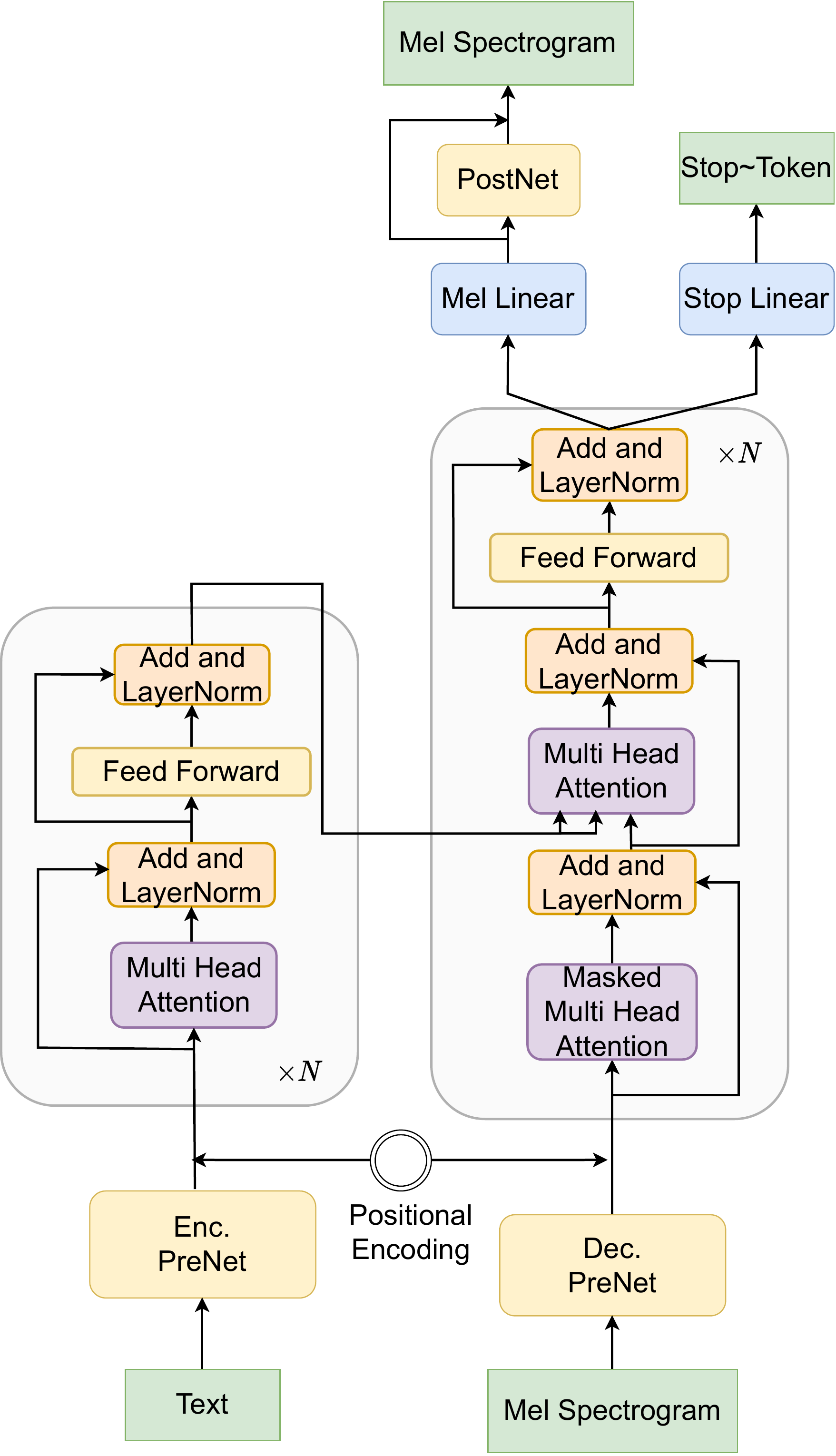}
    \caption{The overview of the Transformer based Text-to-speech framework. Adapters are added according to Fig \ref{fig:adaptarchi} in every transformer layer.}
    \label{TransformerTTS}
\end{figure}

The base model of TTS in this paper is based on \cite{li2019neural}. The overall architecture is shown in Figure \ref{TransformerTTS}. For more details, interested readers can refer to \cite{li2019neural}. In this work, speaker representation used during training the TTS model is computed using an algorithm described in \cite{li2017deep}. The dimension of speaker embeddings is $512$. The output mel-spectogram is converted into audio via a pre-trained HiFiGAN \cite{kong2020hifi} based vocoder. The other necessary  hyperparameters required for training and preprocessing the TTS is given in Table \ref{preprocessing} and Table \ref{training}.
\renewcommand{\arraystretch}{1.5}
\renewcommand{\arraystretch}{1.5}
\begin{table}[h]
\centering
\resizebox{\columnwidth}{!}{
\begin{tabular}{llllll}
\hline
\multicolumn{2}{c}{Audio} & \multicolumn{2}{c}{STFT} & \multicolumn{2}{c}{MEL} \\ \hline
trim\_top\_db   & 23      & filter\_length   & 1024  & n\_mel\_channels & 80   \\
sampling\_rate  & 22050   & hop\_length      & 256   & mel\_fmin        & 0    \\
max\_wav\_value & 32768.0 & win\_length      & 1024  & mel\_fmax        & 8000 \\ \hline
\end{tabular}}
\caption{Parameters for audio preprocessing parameters required for converting raw audio data into mel-spectrogram.}
\label{preprocessing}
\end{table}
\begin{table}[h]
\large
\centering
\resizebox{\columnwidth}{!}{
\begin{tabular}{cccc}
\hline
\multicolumn{2}{c}{\textbf{Optimizer}}                     & \multicolumn{2}{c}{\textbf{Model}}                     \\ \hline
\textbf{batch\_size}        & 16                           & \textbf{encoder\_layer}                    & 4         \\
\textbf{betas}              & {[}0.9, 0.98{]}              & \textbf{encoder\_head/decoder\_head}       & 2/2       \\
\textbf{grad\_clip\_thresh} & 1.0                          & \textbf{encoder\_hidden/decoder\_hidden}   & 256/256   \\
\textbf{total\_step}        & 900000                       & \textbf{decoder\_layer}                    & 6         \\
\textbf{warm\_up\_step}     & 4000                         & \textbf{conv\_filter\_size}                & 1024      \\
\textbf{anneal\_steps}      & {[}300000, 400000, 500000{]} & \textbf{conv\_kernel\_size}                & {[}9,1{]} \\
\textbf{anneal\_rate}       & 0.3                          & \textbf{encoder\_dropout/decoder\_dropout} & 0.2/0/2   \\ \hline
\end{tabular}}
\caption{Optimizer and model parameters used during training and inference.}
\label{training}
\end{table}
\renewcommand{\arraystretch}{2}
\begin{table*}[t]
\Large
\centering
\resizebox{2\columnwidth}{!}{
\begin{tabular}{lllllllllllllll}
\hline
\multicolumn{1}{c}{\multirow{2}{*}{\textbf{Tasks/Methods}}} & \multicolumn{1}{c}{\multirow{2}{*}{\textbf{Datasets}}} & \multicolumn{2}{c}{\textbf{Finetuning}}                                   & \multicolumn{3}{c}{\textbf{Adapter}}                                                                       & \multicolumn{2}{c}{\textbf{Prefix}}                                       & \multicolumn{3}{c}{\textbf{LoRA}}                                                                          & \multicolumn{3}{c}{\textbf{Convadapter}}                                                                   \\ \cline{3-15} 
\multicolumn{1}{c}{}                                        & \multicolumn{1}{c}{}                                   & \multicolumn{1}{c}{\textbf{lr}} & \multicolumn{1}{c}{\textbf{batch size}} & \multicolumn{1}{c}{\textbf{lr}} & \multicolumn{1}{c}{\textbf{batch size}} & \multicolumn{1}{c}{\textbf{c}} & \multicolumn{1}{c}{\textbf{lr}} & \multicolumn{1}{c}{\textbf{batch size}} & \multicolumn{1}{c}{\textbf{lr}} & \multicolumn{1}{c}{\textbf{batch size}} & \multicolumn{1}{c}{\textbf{r}} & \multicolumn{1}{c}{\textbf{lr}} & \multicolumn{1}{c}{\textbf{batch size}} & \multicolumn{1}{c}{\textbf{c}} \\ \hline
\multirow{2}{*}{SR}                                         & ESD                                                    & 2e-4                            & 64                                      & 2e-2                            & 64                                      & 2                              & 2e-2                            & 64                                      & 2e-4                            & 64                                      & 8                              & 2e-4                            & 64                                      & 2                              \\
                                                            & VCTK                                                   & 8e-5                            & 64                                      & 2e-5                            & 64                                      & 2                              & 2e-3                            & 64                                      & 2e-4                            & 64                                      & 8                              & 2e-4                            & 64                                      & 2                              \\ \hline
\multirow{2}{*}{SER}                                        & ESD                                                    & 2e-4                            & 64                                      & 2e-2                            & 64                                      & 2                              & 2e-2                            & 64                                      & 2e-4                            & 64                                      & 8                              & 2e-3                            & 64                                      & 2                              \\
                                                            & MELD                                                   & 7e-6                            & 64                                      & 8e-6                            & 64                                      & 2                              & 2e-2                            & 64                                      & 8e-4                            & 64                                      & 8                              & 2e-4                            & 64                                      & 2                              \\ \hline
\multirow{3}{*}{ASR}                                        & ESD                                                    & 8e-6                            & 16                                      & 2e-5                            & 16                                      & 2                              & 2e-2                            & 16                                      & 2e-3                            & 16                                      & 8                              & 2e-4                            & 16                                      & 2                              \\
                                                            & Fleurs                                                 & 1e-4                            & 32                                      & 1e-4                            & 32                                      & 2                              & 2e-3                            & 32                                      & 2e-3                            & 32                                      & 8                              & 5e-3                            & 32                                      & 2                              \\
                                                            & Librispeech                                            & 3e-5                            & 64                                      & 1e-4                            & 64                                      & 2                              & 3e-4                            & 64                                      & 3e-4                            & 64                                      & 8                              & 3e-4                            & 64                                      & 2                              \\ \hline
PR                                                          & Librispeech                                            & 3e-4                            & 16                                      & 3e-4                            & 16                                      & 2                              & 3e-4                            & 16                                      & 2e-2                            & 16                                      & 8                              & 2e-4                            & 16                                      & 2                              \\ \hline
KS                                                          & Speech Commands                                        & 8e-6                            & 64                                      & 8e-6                            & 64                                      & 2                              & 2e-4                            & 64                                      & 8e-6                            & 64                                      & 8                              & 2e-4                            & 64                                      & 2                              \\ \hline
IC                                                          & Fluent Commands                                        & 2e-4                            & 8                                       & 2e-4                            & 8                                       & 2                              & 2e-4                            & 8                                       & 2e-4                            & 8                                       & 8                              & 2e-4                            & 8                                       & 2                              \\ \hline
SF                                                          & SNIPS                                                  & 2e-4                            & 8                                       & 2e-4                            & 8                                       & 2                              & 2e-3                            & 8                                       & 2e-3                            & 8                                       & 8                              & 2e-4                            & 8                                       & 2                              \\ \hline
\multirow{2}{*}{TTS}                                        & LibriTTS                                               & 1.7e-6                          & 16                                      & 1.7e-6                          & 16                                      & 2                              & 1.7e-6                          & 16                                      & 1.7e-6                          & 16                                      & 8                              & 1.7e-6                          & 16                                      & 2                              \\
                                                            & L2Arctic                                               & 1.7e-6                          & 16                                      & 1.7e-6                          & 16                                      & 2                              & 1.7e-6                          & 16                                      & 1.7e-6                          & 16                                      & 8                              & 1.7e-6                          & 16                                      & 2                              \\ \hline
\end{tabular}}
\caption{Details of hyperparameters for different tasks used during training. Where lr, c and r represents learning rate, compression ratio and rank respectively. }
\label{hyperparam}
\end{table*}
\subsection{Hyperparameters}
In this section, we present the hyperparameters used for generating the results in this paper for different tasks. We experiment with different learning rate for each task and report the value that produce the best results for each algorithm. We do not explore the hyperparameters for Prefix tuning and LoRA and used the default configuration as mentioned in \cite{pfeiffer2020AdapterHub}. In future work, different configuration for these adapters can be explored. Table \ref{hyperparam}, outline all the parameters for different speech processing tasks.

\end{document}